\pdfoutput=1

\documentclass[11pt]{article}

\usepackage[final]{acl}

\usepackage{times}
\usepackage{latexsym}
\usepackage{array}
\usepackage{longtable}
\usepackage[most]{tcolorbox}
\usepackage[T1]{fontenc}

\usepackage[utf8]{inputenc}

\usepackage{microtype}

\usepackage{inconsolata}

\usepackage{graphicx}

\title{PFDial: A Structured Dialogue Instruction Fine-tuning Method \\ Based on UML Flowcharts}

\author{\normalsize \textbf{Ming Zhang}$^{1}$\thanks{\hspace{1mm} Equal Contribution.}\textbf{,} \ \
        \textbf{Yuhui Wang}$^{1*}$\textbf{,} \ \
        \textbf{Yujiong Shen}$^{1*}$\textbf{,} \ \
        \textbf{Tingyi Yang}$^{1}$\textbf{,} \ \
        \textbf{Changhao Jiang}$^{1}$\textbf{,} \\
        \normalsize \textbf{Yilong Wu}$^{1}$\textbf{,}\ \
        \textbf{Shihan Dou}$^{1}$\textbf{,} \ \
        \textbf{Qinhao Chen}$^{5}$\textbf{,} \ \
        \textbf{Zhiheng Xi}$^{1}$\textbf{,} \ \
        \textbf{Zhihao Zhang}$^{1}$\textbf{,} \ \
        \textbf{Yi Dong}$^{1}$\textbf{,} \\
        \normalsize \textbf{Zhen Wang}$^{2}$\textbf{,} \ \
        \textbf{Zhihui Fei}$^{2}$\textbf{,} \ \
        \textbf{Mingyang Wan}$^{2}$\textbf{,} \ \
        \textbf{Tao Liang}$^{2}$\textbf{,} \\
        \normalsize \textbf{Guojun Ma}$^{2}$\thanks{\hspace{1mm} Corresponding Author.}\textbf{,} \ \
        \textbf{Qi Zhang}$^{1,4\dagger}$\textbf{,} \ \
        \textbf{Tao Gui}$^{3,4}$\textbf{,} \ \
        \textbf{Xuanjing Huang}$^{1,3,4}$ \\
  {$^1$  \normalsize College of Computer Science and Artificial Intelligence, Fudan University} \\
  {$^2$  \normalsize Douyin Co., Ltd.}\\
  {$^3$  \normalsize Institute of Modern Languages and Linguistics, Fudan University}\\
  {$^4$  \normalsize Institute of Trustworthy Embodied Artificial Intelligence, Fudan University}\\
  {$^5$  \normalsize Graduate School of Arts and Sciences, Columbia University}\\
  \texttt{\normalsize mingzhang23@m.fudan.edu.cn}\\
  \texttt{\normalsize maguojun@bytedance.com}\\
  \texttt{\normalsize qz@fudan.edu.cn}\\
}
\usepackage{booktabs}  
\usepackage{tabularx}  
\usepackage{algorithm}
\usepackage{algpseudocode}
\usepackage{amsmath}
\usepackage{multirow}
\usepackage{booktabs} 
\usepackage{xcolor}
\usepackage{colortbl}
\usepackage[most]{tcolorbox}
\usepackage[inkscapelatex=false]{svg}

\begin{document}
\maketitle
\begin{abstract}
Process-driven dialogue systems, which operate under strict predefined process constraints, are essential in customer service and equipment maintenance scenarios. Although Large Language Models (LLMs) have shown remarkable progress in dialogue and reasoning, they still struggle to solve these strictly constrained dialogue tasks. To address this challenge, we construct \textbf{P}rocess \textbf{F}low \textbf{Dial}ogue (\textbf{PFDial}) dataset, which contains 12,705 high-quality Chinese dialogue instructions derived from 440 flowcharts containing 5,055 process nodes. Based on PlantUML specification, each UML flowchart is converted into atomic dialogue units i.e., structured five-tuples. Experimental results demonstrate that a 7B model trained with merely 800 samples, and a 0.5B model trained on total data both can surpass 90\% accuracy. Additionally, the 8B model can surpass GPT-4o up to 43.88\% with an average of 11.00\%. We further evaluate models' performance on challenging backward transitions in process flows and conduct an in-depth analysis of various dataset formats to reveal their impact on model performance in handling decision and sequential branches. The data is released in~\url{https://github.com/KongLongGeFDU/PFDial}.

\end{abstract}

\begin{figure}[t]
    \centering
    \includegraphics[scale=0.5]{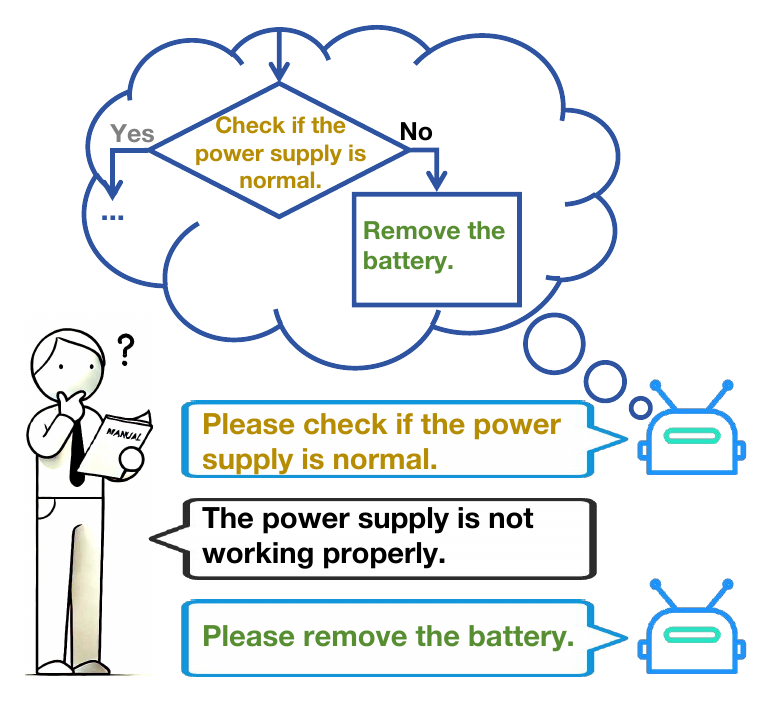}
    \caption{In this scenario, the system interacts with the user based on a flowchart that checks whether the power supply is functioning properly. If the power supply is faulty, the system guides the user to remove the battery. This interaction follows the decision-making process outlined in the flowchart.}
    \label{fig:intro}
\end{figure}

\section{Introduction}
\label{sec:intro}
Process-driven dialogue systems \cite{TOD-Survey}, as a special type of task-oriented dialogue systems, play a crucial role in various real-world applications, particularly in scenarios such as customer service, equipment maintenance, and medical consultation, where strict adherence to predefined process constraints is essential. In these contexts, dialogue systems must navigate through complex decision trees while maintaining precise control over the conversation flow, ensuring both compliance with established procedures and effective user interaction. These process flows typically contain two types of branch: sequential branches that follow a linear progression and decision branches that require conditional routing based on user input. Figure \ref{fig:intro} illustrates an example scenario of process-driven dialogue systems.

The emergence of Large Language Models (LLMs) has brought unprecedented capabilities in natural language understanding and generation, demonstrating remarkable performance in both dialogue and reasoning tasks. These advances suggest potential solutions for process-driven dialogue systems \cite{TOD-Survey, SGP-TOD, TOD-BERT}. However, our empirical evaluation reveals that even state-of-the-art (SOTA) LLMs such as GPT-4o \cite{GPT4} struggle to consistently maintain process constraints while engaging in dialogue. Specifically, these models often deviate from predefined process constraints, make incorrect state transitions, or fail to properly handle complex decision branches, highlighting the need for more specialized solutions.

To address this challenge, we construct \textbf{P}rocess \textbf{F}low \textbf{Dial}ogue (PFDial) dataset, which contains 12,705 high-quality dialogue instructions derived from 440 flowcharts containing 5,055 process nodes. Based on PlantUML specification, each UML flowchart is converted into atomic dialogue units, forming structured five-tuples (flowchart description, current state, user input, next state, robot output). This structured representation enables models to learn precise state transitions while maintaining natural dialogue capabilities. Through supervised fine-tuning (SFT) on PFDial, models can acquire strong controlled reasoning capabilities for process flows effectively following the prescribed state transitions and decision logic.

We conducted comprehensive experiments to address four key questions:

\textbf{(1) How does our approach perform compared to SOTA LLMs?}  
Our main experimental results demonstrate that models with varying parameter sizes can achieve excellent results after SFT on total data of PFDial. For instance, even a 0.5B model can achieve accuracy of 98.99\% and 92.79\% on in-domain and out-of-domain tests, respectively. More impressively, an 8B model achieves 97.02\% accuracy on out-of-domain tests, surpassing GPT-4o by 11.00\%, with over 43.88\% improvement in decision branches.

\textbf{(2) How does model performance scale with training data size?}  
Our data scaling experimental results show that a 7B model can surpass 90\% accuracy with only 800 training samples. As the training dataset increases, the overall performance of the model continues to improve. This underscores PFDial's exceptional effectiveness in enhancing the model's capability for controlled reasoning tasks with minimal data.

\textbf{(3) How effective is our approach in handling backward transitions?}  
We evaluated the model's performance on more challenging backward transitions in decision branches using our constructed dataset, PFDial-H. This specialized benchmark focuses on cases where the next state returns to a previous point in the process flow. These transitions are particularly challenging due to their relative scarcity and the complex reasoning they require. Results on PFDial-H further validate our approach's superiority in challenging controlled reasoning tasks.

\textbf{(4) How do different state representation formats affect model performance?}  
Through systematic analysis of three different dataset formats, we provide interpretable insights into how different formats impact model performance on decision and sequential branches, offering valuable guidance for future research.

Overall, our contributions can be summarized as follows:
\begin{itemize}
    \item We have developed the Process Flow Dialogue (PFDial) dataset, which is derived from 440 flowcharts encompassing 5,055 process nodes. This dataset contains 12,705 high-quality dialogue instructions, serving as a valuable resource for training process-driven dialogue systems.
    \item The comprehensive experiments demonstrate that the PFDial dataset is highly effective. Even models with a smaller number of parameters (e.g., 0.5B, 1B, 1.5B) or those trained on relatively limited data (800 training samples) can achieve high accuracy. An 8B model achieves 97.02\% accuracy on out-of-domain tests, surpassing GPT-4o by 11.00\%, with over 43.88\% improvement in decision branches.
    \item We demonstrate our model's superior performance on complex backward transitions in decision branches using the PFDial-H benchmark, highlighting its capability in handling rare and intricate reasoning tasks. Additionally, we provide insights into the impact of different dataset formats on model performance, offering guidance for future research.
\end{itemize}

\begin{table}[t]
  \small
  \setlength{\tabcolsep}{4pt}  
  \begin{tabularx}{\columnwidth}{Xrrr}  
  \toprule
  \textbf{Statistics} & \textbf{Train} & \textbf{ID Test} & \textbf{OOD Test} \\
  \midrule
  Flowcharts & $440$ & $80$ & $80$ \\
  State Nodes & $5055$ & $902$ & $1262$ \\
  Sequential Samples & $9029$ & $1628$ & $2265$ \\
  Decision Samples & $3676$ & $645$ & $698$ \\  
  Dialogue Samples & $12705$ & $2273$ & $2963$ \\
  Avg. Length & $277.16$ & $270.57$ & $326.10$ \\
  \bottomrule
  \end{tabularx}
  \caption{Statistics of the PFDial Dataset}
  \label{tab:dataset_statistics}
\end{table}

\begin{figure*}[t]
    \centering
    \includegraphics[width=0.95\textwidth]{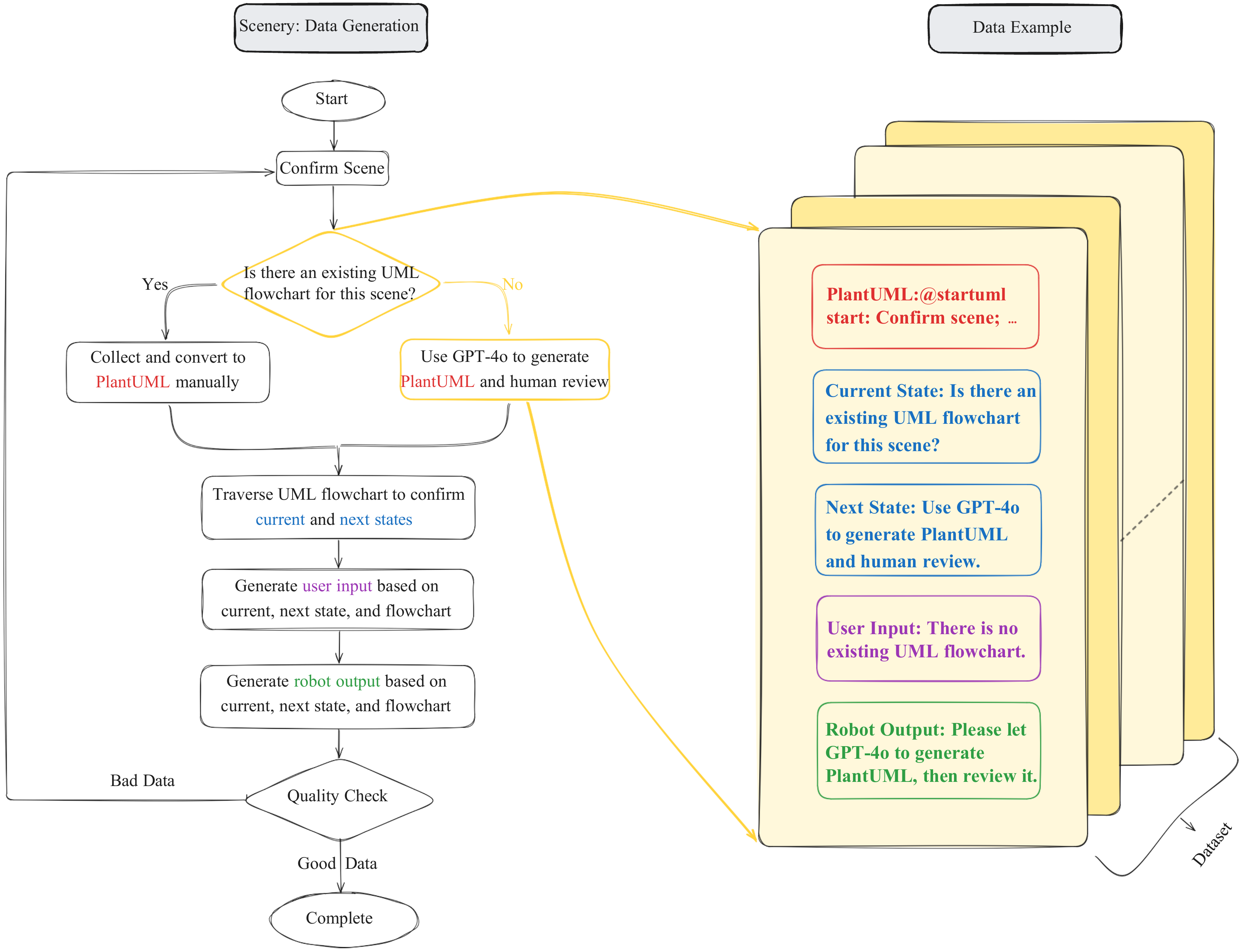}
    \caption{The left side illustrates the data construction process.
The right side shows an example of the five-tuple dataset generated based on the leftside flowchart.}
    \label{fig:dataset}
\end{figure*}

\section{PFDial}
In this section, we introduce PFDial, a Chinese dataset specifically designed for process-driven dialogue systems.

\subsection{Dataset Overview}
\paragraph{PFDial}
The construction of PFDial follows a systematic pipeline, including flow chart collection, textual representation conversion, state transition Information extraction, prompt generation, and data validation. The dataset combines real-world scenarios and synthetic data, achieving broad coverage of practical applications while maintaining high quality and diversity. Table \ref{tab:dataset_statistics} presents detailed statistics of the PFDial dataset.

\paragraph{PFDial-H}
Considering the prevalence and importance of backward transitions in practical applications, we specifically constructed a supplementary dataset that incorporates backward transition mechanisms called PFDial-Hard (PFDial-H). By strategically adding backward transition nodes to existing flowcharts using GPT-4o, we implemented backward transition functionality for cases where conditions are not met. This improvement was applied to both the out-of-domain test and training flowcharts, generating new training samples through the same prompt augmentation process. Table \ref{tab:pfd-h} in Appendix \ref{appendix:pfdiag-h} presents detailed statistics of the PFDial-H dataset.

\subsection{Dataset Construction Process}
\paragraph{Flow Chart Collection}

Through extensive research, we identified and categorized 90 specific business scenarios, details of which can be found in the Appendix \ref{tab:Business Scenarios}. Additionally, we designed a carefully constructed template dataset. After identifying the business scenarios, we collected the flowcharts manually or automatically, depending on whether pre-existing flowcharts were available. This process, combining automation with human review, significantly improved the efficiency and accuracy of UML flowchart generation.

\paragraph{Textual Representation Conversion}

To efficiently convert flowcharts into machine-readable formats, we conducted a comparative study of several text-based representation schemes, including PlantUML\footnote{\url{https://plantuml.com/}}, chart-mage\footnote{\url{https://chartmage.com/intro.html}}, nomnoml\footnote{\url{https://www.nomnoml.com/}}, and Mermaid\footnote{\url{https://mermaid.js.org/}}. After comprehensive evaluation, we selected PlantUML as the standard format. This decision was made for several reasons: First, PlantUML employs a code-like structured representation utilizing syntax features such as indentation, branching, and loops, making flowchart descriptions both intuitive to read and convenient for program processing. Second, a preliminary experiment using GPT-4o demonstrated that the PlantUML format exhibited superior accuracy compared to other approaches. Details can be find in Table \ref{tab:different-flowchart-formats} in Appendix \ref{appentix:different-flowchart-format}. During data processing, we represented all flowcharts in PlantUML format and generated standardized state nodes and their transition relationships through parsing, providing a unified representation for subsequent five-tuple dataset generation.

\paragraph{State Transition Information Extraction}
Based on the standardized PlantUML representation, we developed a specialized algorithm to extract complete state transition information, which is shown in Algorithm \ref{alg:parsePlantUML} in Appendix \ref{appendix:algorithm}. During this process, we got all existing paths and identified 5,055 distinct state nodes from the training set. Each state transition pair (current state -> next state) strictly corresponds to a specific path in the flowchart, ensuring the accuracy and consistency of the state information extraction.

\paragraph{Prompt Generation}
To create quality training samples, we used the GPT-4o model for bidirectional prompt augmentation for each state transition. This involved generating user input and robot output based on the current and next states, along with the flowchart. Detailed prompts can be found in Appendix \ref{appentix:Prompt}.

\begin{table*}[t]
    \small
    \centering
    \resizebox{0.96\textwidth}{!}{
    \newcolumntype{L}{>{\raggedright\arraybackslash}p{3.5cm}} 
    \newcolumntype{C}{>{\centering\arraybackslash}p{1.5cm}} 
    \begin{tabular}{L|C C C|C C C}  
    \toprule
    \multirow{2}{*}{\parbox{3.5cm}{\centering \textbf{Model}}} & \multicolumn{3}{c|}{\textbf{ID-test}} & \multicolumn{3}{c}{\textbf{OOD-test}} \\
    \cmidrule(lr){2-4} \cmidrule(lr){5-7}
    & \textbf{Acc} & \textbf{Decision Acc} & \textbf{Sequential Acc} 
    & \textbf{Acc} & \textbf{Decision Acc} & \textbf{Sequential Acc} \\
    \midrule
    \multicolumn{7}{l}{\textbf{Baselines}} \\
    LLaMA-3.1-8B-Instruct & $-$ & $-$ & $-$ & $13.20$ & $2.16$ & $15.00$ \\
    Claude-3.5-Sonnet & $-$ & $-$ & $-$ & $62.74$ & $22.06$ & $69.40$ \\
    Qwen2.5-7B-Instruct & $-$ & $-$ & $-$ & $65.87$ & $37.88$ & $71.34$ \\
    Gemini-2.0-Flash-Exp & $-$ & $-$ & $-$ & $75.17$ & $47.48$ & $79.66$ \\
    DeepSeek-v3 & $-$ & $-$ & $-$ & $79.02$ & $47.72$ & $84.11$ \\
    GPT-3.5-Turbo & $-$ & $-$ & $-$ & $79.76$ & $39.57$ & $86.29$ \\
    GPT-4o & $-$ & $-$ & $-$ & $86.29$ & $51.80$ & $91.90$ \\
    \midrule
    \multicolumn{7}{l}{\textbf{FineTuned on PFDial}} \\
    LLaMA-3.2-1B & $98.90$ & $\textbf{98.59}$ & $98.96$ & $93.57$ & $87.05$ & $94.62$ \\
    LLaMA-3.2-3B & $98.77$ & $98.02$ & $98.91$ & $95.81$ & $91.37$ & $96.53$ \\
    LLaMA-3.1-8B & $\textbf{99.03}$ & $98.31$ & $\textbf{99.17}$ & $\textbf{97.29}$ & $\textbf{96.88}$ & $97.35$ \\
    Qwen2.5-0.5B & $98.99$ & $98.02$ & $\textbf{99.17}$ & $91.35$ & $89.45$ & $91.66$ \\
    Qwen2.5-1.5B & $98.90$ & $97.46$ & $\textbf{99.17}$ & $94.00$ & $88.97$ & $94.82$ \\
    Qwen2.5-3B & $98.77$ & $98.31$ & $98.85$ & $94.97$ & $89.69$ & $95.84$ \\
    Qwen2.5-7B & $98.94$ & $98.02$ & $99.11$ & $96.51$ & $90.65$ & $\textbf{97.47}$ \\
    \bottomrule
    \end{tabular}
    }
    \caption{Results on PFDial: Decision Acc represents the accuracy of the decision branch, and Sequential Acc reflects the accuracy of the sequential branch.  Acc is The overall accuracy.}
    \label{tab:base_model_performance}
\end{table*}

  \begin{figure*}[tb]
    \begin{center}
        \begin{minipage}{0.48\textwidth}  
            \centering
            \includegraphics[width=\textwidth]{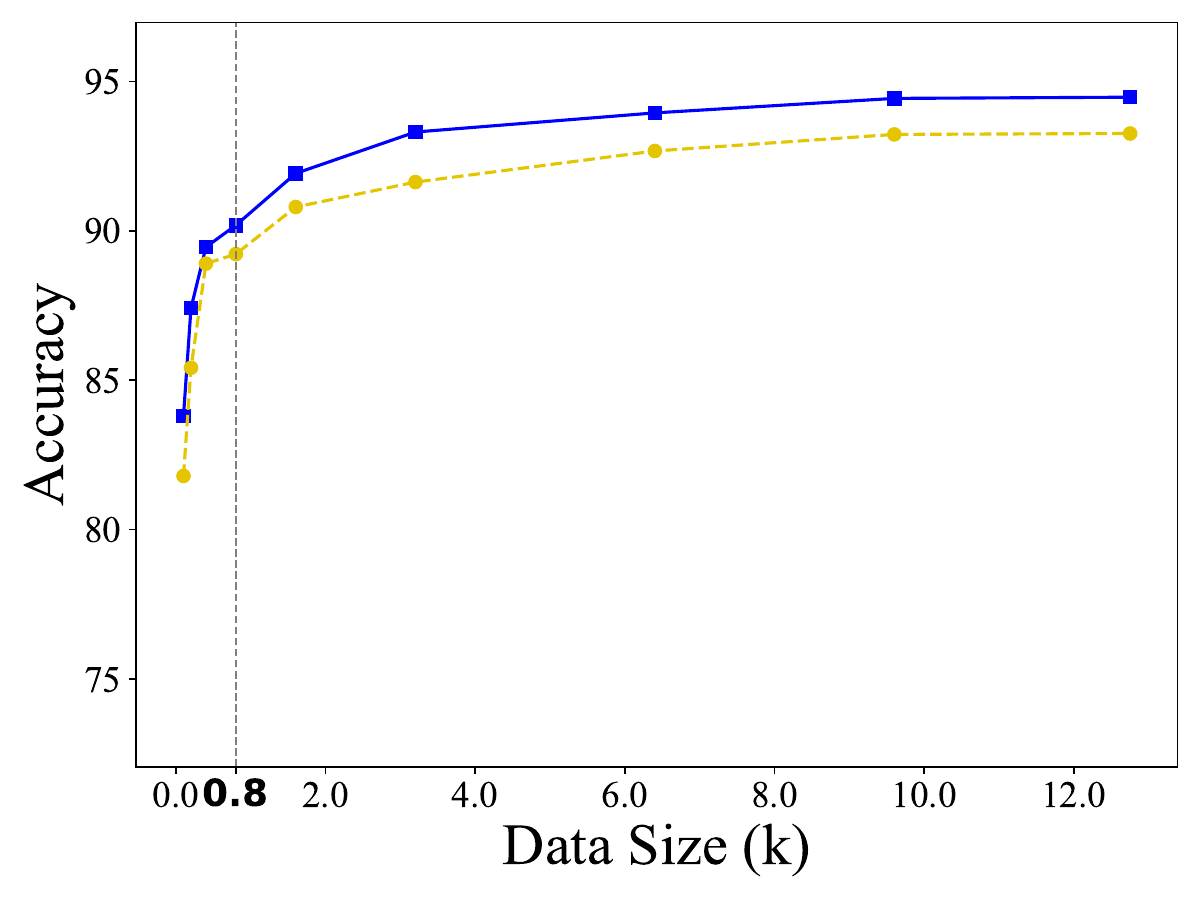}
        \end{minipage}
        \hspace{0.02\textwidth} 
        \begin{minipage}{0.48\textwidth}  
            \centering
            \includegraphics[width=\textwidth]{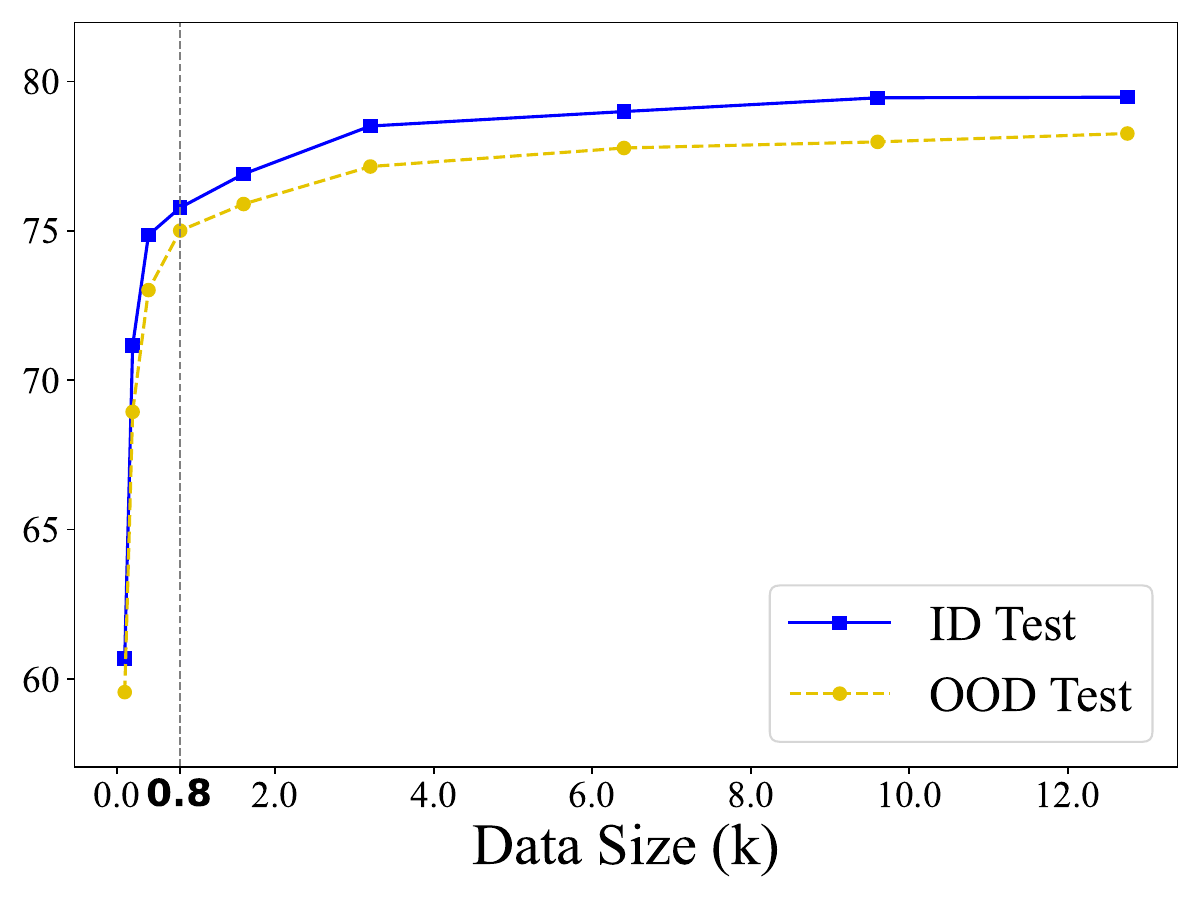}
        \end{minipage}
        \caption{Results on Data Scaling: The left plot shows the accuracy of ID and OOD tests using data scaling strategy \textbf{(a)}, while the right plot shows the accuracy using data scaling strategy \textbf{(b)}.}
    \label{scaling_figures}
    \end{center}
    \end{figure*}
    
\paragraph{Data Validation}
To ensure the reliability of the dataset, we implemented a rigorous multi-level validation process: first, ensuring that all state nodes strictly correspond to the original flowcharts; second, validating the syntax correctness of the PlantUML; and finally, checking the logical consistency between user inputs and state transitions. Any data that does not meet these criteria will be regenerated.

This dataset construction methodology not only ensures data quality and diversity, but also provides a solid foundation for subsequent model training. Through systematic construction processes and strict quality control, the PFDial dataset effectively balances authenticity, standardization, and scalability requirements.
Figure \ref{fig:dataset} illustrates the data construction process flowchart and an example of the five-tuple dataset generated based on the data construction flowchart.

\section{Experiments}

\begin{table*}[t]
    \small
    \centering
    \resizebox{0.96\textwidth}{!}{
    \newcolumntype{L}{>{\raggedright\arraybackslash}p{3.5cm}}
    \newcolumntype{C}{>{\centering\arraybackslash}p{1.5cm}}
    \newcolumntype{Y}{>{\centering\arraybackslash}p{2cm}}
    \begin{tabular}{L|C Y Y|C C C}  
    \toprule
    \multirow{2}{*}{\parbox{3.5cm}{\centering \textbf{Model}}} & \multicolumn{3}{c|}{\textbf{PFDial-H}} & \multicolumn{3}{c}{\textbf{OOD-test of PFDial}} \\
    \cmidrule(lr){2-4} \cmidrule(lr){5-7}
    & \textbf{Acc} & \textbf{Backward Acc(Dist $<$5)} & \textbf{Backward Acc(Dist $\geq$5)} 
    & \textbf{Acc} & \textbf{Decision Acc} & \textbf{Sequential Acc} \\
    \midrule
    \multicolumn{7}{l}{\textbf{Baselines}} \\
    LLaMA-3.1-8B-Instruct & $15.00$ & $13.64$ & $15.52$ & $13.20$ & $2.16$ & $15.00$ \\
    Claude-3.5-Sonnet & $57.50$ & $59.09$ & $56.90$ & $62.74$ & $22.06$ & $69.40$ \\
    Qwen2.5-7B-Instruct & $31.25$ & $31.82$ & $31.03$ & $65.87$ & $37.88$ & $71.34$ \\
    Gemini-2.0-Flash-Exp & $26.25$ & $22.73$ & $27.59$ & $75.17$ & $47.48$ & $79.66$ \\
    DeepSeek-v3 & $40.00$ & $31.82$ & $43.10$ & $79.02$ & $47.72$ & $84.11$ \\
    GPT-3.5-Turbo & $51.25$ & $54.55$ & $50.00$ & $79.76$ & $39.57$ & $86.29$ \\
    GPT-4o & $58.75$ & $68.18$ & $55.17$ & $86.29$ & $51.80$ & $91.90$ \\
    \midrule
    \multicolumn{7}{l}{\textbf{Secondary Fine-tuning}} \\
    Qwen2.5-0.5B  & $58.75$  & $77.27$  & $51.72$  & $50.77$  & $52.67$  & $39.09$  \\
    Qwen2.5-1.5B  & $47.50$  & $45.45$  & $48.28$  & $87.90$  & $88.82$  & $82.25$  \\
    Qwen2.5-3B    & $57.50$  & $59.09$  & $56.90$  & $79.73$  & $80.91$  & $72.42$  \\
    Qwen2.5-7B    & $66.25$  & $90.91$  & $56.90$  & $76.04$  & $76.82$  & $71.22$  \\
    \midrule
    \multicolumn{7}{l}{\textbf{Integrated Training}} \\
    Qwen2.5-0.5B  & $65.00$  & $72.73$  & $62.07$  & $92.76$  & $93.22$  & $89.93$  \\
    Qwen2.5-1.5B  & $70.00$  & $72.73$  & $68.97$  & $93.87$  & $94.82$  & $88.01$  \\
    Qwen2.5-3B    & $75.00$  & $\mathbf{90.91}$  & $68.97$  & $94.57$  & $96.18$  & $84.65$  \\
    Qwen2.5-7B    & $\mathbf{76.25}$  & $86.36$  & $\mathbf{72.41}$  & $\mathbf{96.31}$  & $\mathbf{96.96}$  & $\mathbf{92.33}$  \\
    \bottomrule
    \end{tabular}
    }
    \caption{Results on PFDial and PFDial-H: Backward Acc represents the accuracy of backward transition.}
    \label{tab:model_performance on PFDial-H}
\end{table*}

In this chapter, we present a series of comprehensive experiments to address the four key questions mentioned in Section \ref{sec:intro}. We conducted the main experiment, data scaling experiment, backward transition studies, and format ablation studies respectively.

\subsection{Experimental Setup}

\paragraph{Base Models} We evaluate our method using two series of base models with varying parameter sizes. The first series includes Qwen2.5 models \cite{Qwen2.5} ranging from 0.5B to 7B parameters (Qwen2.5-0.5B, Qwen2.5-1.5B, Qwen2.5-3B, and Qwen2.5-7B). The second series consists of Llama3 models \cite{Llama} spanning from 1B to 8B parameters (Llama3.2-1B, Llama3.2-3B, and Llama3.1-8B). This diverse selection of models enables us to comprehensively analyze the impact of model scale on performance.

\paragraph{Baselines} For comparison, we select a comprehensive set of both open-source and proprietary state-of-the-art (SOTA) LLMs that have demonstrated strong performance across various NLP tasks. These include proprietary models like GPT-4o \cite{GPT4}, GPT-3.5-turbo \cite{GPT3.5}, Gemini-2.0-flash-exp \cite{Gemini}, and Claude-3.5-sonnet \footnote{\url{https://www.anthropic.com/news/claude-3-family}}, as well as open-source models such as DeepSeek-v3 \cite{DeepSeek-V3}, Llama3.1-8b-instruct \cite{Llama}, and Qwen2.5-7b-instruct \cite{Qwen2.5}. These models represent the current frontier of language model capabilities and serve as strong baselines for evaluating our approach.

\paragraph{Datasets}
Our main experiments utilize the PFDial dataset containing 12,705 training samples. For evaluation, we maintain two test sets: our out-of-domain test set comprising 698 decision branches and 2,963 sequential branches, and an in-domain test set containing 645 decision branches and 1628 sequential branches. For backward transition experiments, we use the specially constructed PFDial-H dataset. For format ablation studies, we construct corresponding training and test sets in three different state representation formats to systematically evaluate their impact on model performance.

\paragraph{Implementation Details} 
We combine the PFDial dataset with the general dialogue dataset BELLE \cite{BELLE, wen2023chathome} in a 1:1 ratio for SFT. For the training process, we utilize the OpenRLHF\cite{OpenRLHF} framework. For instance, the training of Qwen2.5-7B model is conducted on 8 H20 GPUs, with a total training time of approximately 1 hour. For detailed hyper-parameter configurations, please refer to Table \ref{tab:hyperparams} in Appendix \ref{appendix:implementation_details}.

\paragraph{Evaluation Metrics}
Model performance is evaluated by exact match accuracy between predicted and ground truth next states. For a prediction to be considered correct, it must exactly match the ground truth state transition. 
Specifically, our dataset contains two types of samples: sequential samples and decision samples. Sequential samples refer to cases where there is only one unique next state given the current state. Decision samples refer to cases where there are at least two possible next states for the current state.  Accordingly, our accuracy can be further refined into two types: sequential accuracy and decision accuracy.



\subsection{Main Results}  Table \ref{tab:base_model_performance} presents our detailed experimental results on the main experiments.
Our experiments demonstrate significant improvements over baseline models across all parameter scales. Even our smallest 0.5B parameter model achieves 91.35\% accuracy on out-of-domain tests, with particularly strong performance on decision branches. Our 8B model achieves sota performance with 97.29\% accuracy on out-of-domain tests, surpassing GPT-4o by 11\%. In decision branches, our 8B parameter model achieves 96.88\% accuracy, surpassing GPT-4o by 43.88\%. See Appendix \ref{case study} for a detailed case study analyzing the reasons behind model errors.

\begin{figure*}[tb]
\begin{center}
    \begin{minipage}{0.48\textwidth}  
        \centering
        \includegraphics[width=\textwidth]{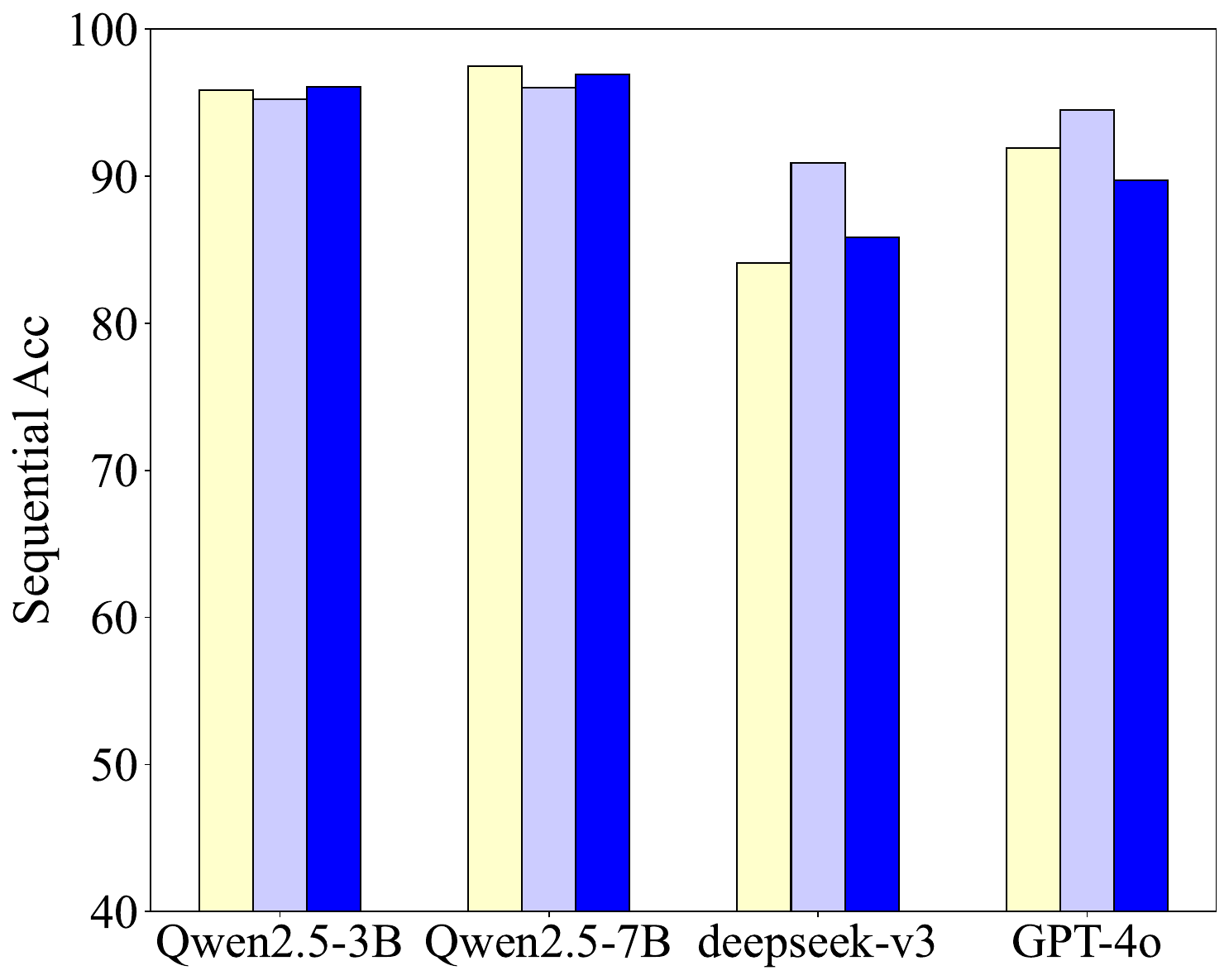} \\
    \end{minipage}
    \hspace{0.02\textwidth} 
    \begin{minipage}{0.48\textwidth}  
        \centering
        \includegraphics[width=\textwidth]{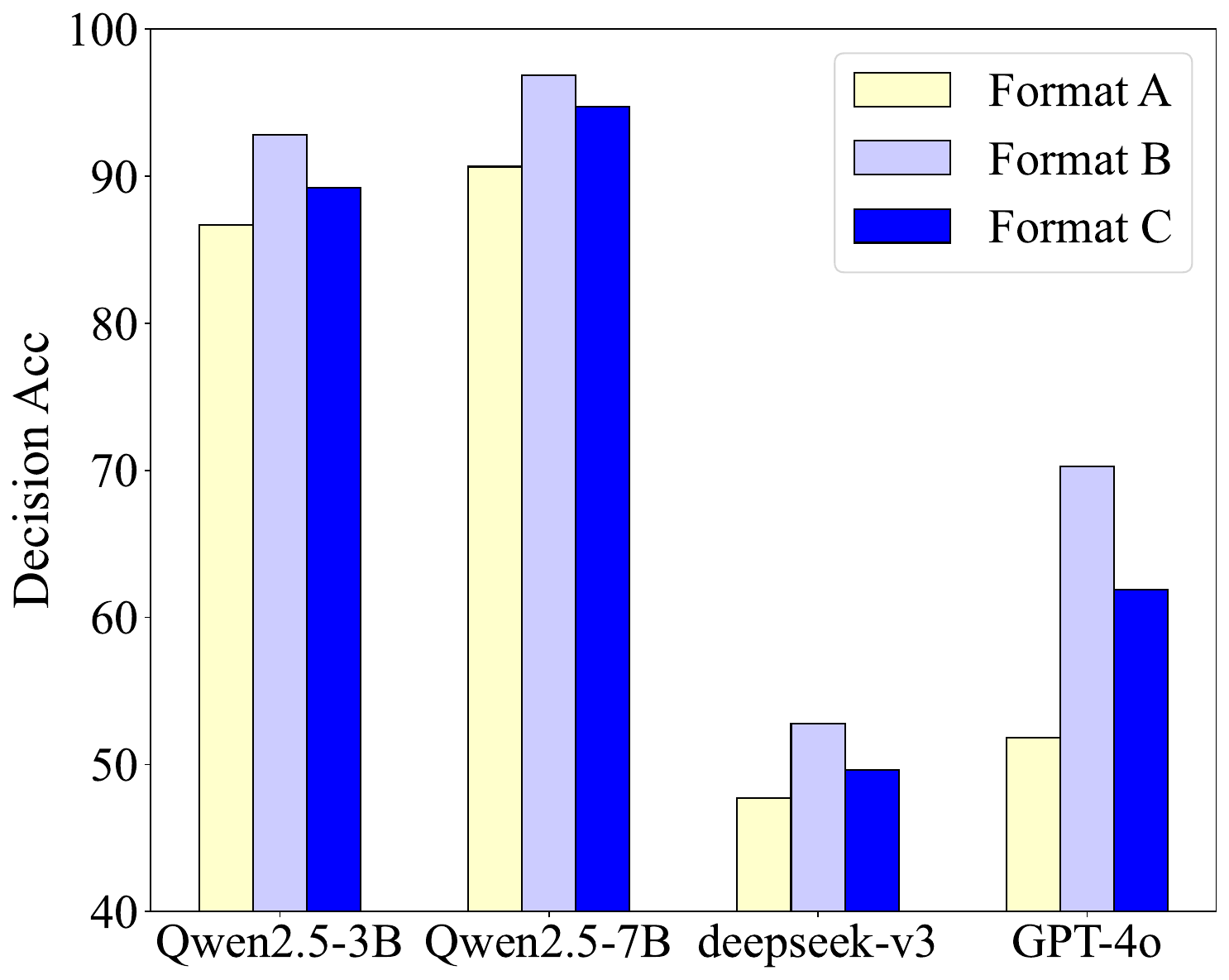} \\
    \end{minipage}
    \caption{Results on Model Performance with Different Formats: The left plot shows the sequential accuracy for different models across three different formats (Format-NL, Format-SC, and Format-Hybrid). The right plot shows the decision accuracy for the same models under the same formats. }
\label{format_figures}
\end{center}
\end{figure*}
\begin{figure*}[t]
\vskip 0.2in
\begin{center}
    \includegraphics[width=\textwidth]{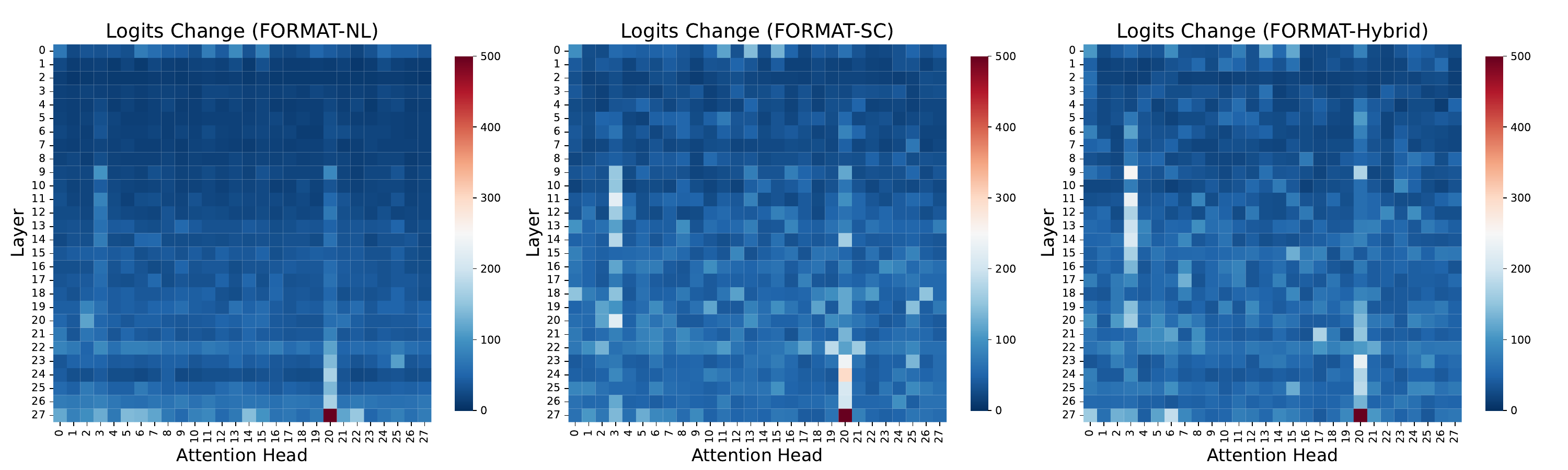} \\
    \caption{Logits changes of Three formats}
    \label{fig:logits_change}
\end{center}
\vskip -0.1in
\end{figure*}

\subsection{Data Scaling Experiments} 
\label{sec:scale_experiment}
\paragraph{Experimental Setup} To investigate data efficiency, we conducted experiments with varying training data sizes from 100 to 12,705 samples. We employed two data scaling strategies: \textbf{(a)} keeping fixed 12,000 general dialogue data samples while gradually increasing PFDial data, and \textbf{(b)} maintaining a 1:1 ratio mixing of general dialogue data and PFDial data.

\paragraph{Results and Analysis} Figure \ref{scaling_figures} presents our experimental results on the data scaling experiments. The results demonstrate remarkable performance even with limited data: using only 800 samples, a 7B model can surpass 90\% accuracy on OOD test. Performance continues to improve consistently with increased data volume, though we observe diminishing returns after approximately 3,000 samples. The comparable performance across both data scaling strategies validates the effectiveness of our PFDial dataset, demonstrating its robust data efficiency regardless of the mixing approach. For comprehensive results on decision accuracy and sequential accuracy across various data scaling configurations, please refer to Appendix \ref{appendix:scaling}.


\subsection{Backward Transition Studies} 
\paragraph{Experimental Setup} We evaluated models' capability in handling complex backward transitions on the PFDial-H test set, which provides a more rigorous assessment of models' ability to strictly follow process constraints. The details of PFDial-H test set is shown in table \ref{tab:pfd-h} in Appendix \ref{appendix:pfdiag-h}. For handling backward transitions, we compared two approaches: integrated training, which incorporates 440 PFDial-H training data samples to our PFDial training data during initial training, and secondary fine-tuning, which applies a secondary fine-tuning phase using PFDial-H data on the SFT-completed model with 440 PFDial-H training data samples. 

\paragraph{Results and Analysis}
Detailed experimental results are shown in Table \ref{tab:model_performance on PFDial-H}. Our results demonstrate that our method achieves optimal performance in handling backward branches, with the integrated training approach yielding superior results by maintaining robust performance on forward transitions while significantly improving accuracy in backward transition cases. Specifically, with integrated training, the Qwen2.5-7B model achieves 76.25\% overall accuracy on PFDial-H, with 86.36\% accuracy on backward transitions with distance less than 5, and 72.41\% accuracy on transitions with distance greater than or equal to 5. Meanwhile, the model maintains a high accuracy of 96.31\% on the OOD test.

In contrast, secondary fine-tuning not only fails to improve performance on backward transition cases but also reduces performance on the PFDial dataset, with Qwen2.5-7B's OOD test accuracy dropping from 96.31\% to 76.04\%. These results emphasize the importance of integrating backward transition samples during the initial training process rather than treating them as a post-hoc fine-tuning step.

\subsection{Format Abliation Studies} 
\paragraph{Experimental Setup} We conducted experiments for three state representation formats: Format-NL: natural language description (default method), Format-SC: state codes (e.g., S1, C2), and Format-Hybrid: combining codes and descriptions to explore the impact of different data formats on model performance and the underlying reasons. Specific cases of data in different formats and the visualization results can be seen in Appendix \ref{appendix:data and attention in different formats}. To ensure fairness, we fine-tuned the base model on data in all three formats and tested it with corresponding test sets containing the same content.

We then froze the attention heads of each layer and compared the model's output logits with the original logits in these scenarios. By examining the logits changes , we assessed the impact of each attention head after fine-tuning with different formats. Finally, we visualized these attention heads to gain deeper insights into their roles.

\paragraph{Results and Analysis}
The model performance with different formats are shown in Figure \ref{format_figures}. The results indicate that Format B achieved the highest accuracy in most cases, particularly on Decision Accuracy, however, it performed slightly worse than other formats on Sequential Accuracy. The following experimental results, to some extent, shed light on the reasons behind this phenomenon.

The comparison results of the three formats's attention heads are shown in Figure \ref{fig:logits_change}. The models fine-tuned with FormatB and FormatC showed a more uniform and diverse distribution of attention head contributions. This can be explained that both of the latter formats introduced code state identifiers, requiring the language model to learn both the sequence of state transitions and the correspondence between code states and natural language states. Thus more attention heads with different functions were activated.

In particular, $head\_layer_{27}\_head_{20}$ is crucial in all three formats. We visualized this attention head's scores for all three formats in Figure \ref{fig:attention_scores_a}, Figure \ref{fig:attention_scores_b}, and Figure \ref{fig:attention_scores_c}, respectively, focusing on significant differences. The attention scores for both the natural language and hybrid formats were concentrated at the intersection of the user's current state and the corresponding state in the flowchart. Format-Hybrid, which includes some state codes, showed a more moderated concentration. In contrast, Format-SC did not exhibit such clustering. We hypothesize that the introduction of state codes allows the model's attention to generalize across different input parts, rather than being confined to specific segments. This enables models to better understand user inputs and facilitates learning of global logic, such as condition checking and state selection. This also reasonably explains previous results.

\section{Related Work}
\subsection{Controllable Reasoning in LLMs}
Controllable reasoning in LLMs has gained significant attention in recent years. \citet{InstructGPT} pioneered instruction-guided control via Reinforcement Learning from Human Feedback (RLHF), combining supervised fine-tuning (SFT), reward model training, and Proximal Policy Optimization (PPO) to align outputs with human preferences. While effective, this approach requires complex annotation and training \cite{Prefix-Tuning}. In contrast, our approach simplifies the process by encoding reasoning logic into structured UML state flowcharts, guiding learning through SFT alone. This provides a clear, human-readable control mechanism, addressing the 'reasoning opacity' challenge \cite{Controllable-Text-Generation-Survey}.

\subsection{Graph-based Enhanced Reasoning}
Previous research \cite{Roadmap} has explored graph-based approaches to enhance the reasoning capabilities of LLMs. Several works \cite{Controlled-Text-Generation-With_Dynamic-Attribute-Graphs, ChatKBQA, Reasoning-on-Graphs} have used Knowledge Graph (KG) structural information to reduce hallucinations by breaking reasoning into path extraction and inference steps. Similarly, \citet{Enhancing-Logical-Reasoning} showed that graph-based training improves multi-hop reasoning accuracy. However, these methods mainly treat graphs as external knowledge sources rather than explicit control mechanisms.

\subsection{LLM-based Dialogue Systems}
Task-Oriented Dialogue (TOD) systems help users achieve specific goals through conversations \cite{TOD-Survey}. Current approaches fall into two main categories: Pipeline-based Approaches, which separate dialogue systems into modules with LLMs handling specific tasks \cite{ZeroShotBertAdapters, ExploringZero, BertForJoint, CoF-CoT}, and End-to-End Approaches, which use LLMs to generate responses based on the entire dialogue history \cite{SimpleMTOD, TransferTOD, SGP-TOD, TOD-BERT, PromptingTOD}. Pipeline-based Approaches offer better transparency but require extensive annotated data, while End-to-End Approaches are simpler but less controllable and demand higher model capabilities.

\section{Conclusion}

In this paper, we introduce the PFDial dataset, a novel resource designed to enhance process-driven dialogue systems. By utilizing structured dialogue instructions derived from UML flowcharts, PFDial provides a robust framework for training models to handle complex decision-making and sequential processes. Our experiments demonstrate that models fine-tuned on PFDial achieve high accuracy, even with limited training data, and outperform sota LLMs like GPT-4o on specific tasks.

We conducted an in-depth analysis of backward transitions using the PFDial-H dataset, highlighting the importance of integrated training approaches for maintaining strong performance across diverse dialogue scenarios. Additionally, we explored the impact of various data representation formats, finding that structured state codes significantly improve the accuracy of state transition predictions.

Overall, our work underscores the potential of structured datasets like PFDial to advance process-driven dialogue systems, offering new insights into the design and training of models for precise and controlled reasoning. Future research will focus on expanding the dataset to cover more scenarios and refining training methodologies to enhance model generalization and adaptability.

\section*{Limitations}
Our research presents a comprehensive set of experiments, yet it is not without limitations.
First, the Chinese-centric nature of our dataset introduces potential cross-lingual generalization constraints. Second, the scarcity of standardized flowchart benchmarks in Chinese process specifications increases the risk of domain-specific biases, despite our rigorous validation framework. Besides, the potential residual inconsistencies in flow-to-text conversion may emerge from the inherent subjectivity in interpreting semantic structures.

\section*{Acknowledgments}
The authors wish to thank anonymous reviewers for their helpful comments. This work was partially funded by National Natural Science Foundation of China (No.62376061,62206057), Shanghai Rising-Star Program (23QA1400200), Natural Science Foundation of Shanghai (23ZR1403500), Program of Shanghai Academic Research Leader under grant 22XD1401100.

\bibliography{anthology, custom}
\appendix

\section{Algorithm}

\subsection{Parse PlantUML Code}
This algorithm parses the given PlantUML code by line by line. It initializes an empty dictionary, \textit{nodeDict}, to store state nodes. Then, it calls \textsc{parseSequential} (Algorithm \ref{alg:parseSequential}) to process the sequential flow. Finally, it returns all paths originating from the start node, representing the complete execution flow.
\begin{algorithm}
\caption{Parse PlantUML Code}
\label{alg:parsePlantUML}
\begin{algorithmic}[1]
\Function{parsePlantUML}{code}
    \State Split code into lines
    \State Initialize an empty dictionary \textit{nodeDict} to store nodes
    \State Call \textsc{parseSequential}
    \State \Return all paths originating from the start node
\EndFunction
\end{algorithmic}
\end{algorithm}
\label{appendix:algorithm}

\subsection{Parse Sequential Blocks}
This algorithm parses the sequential execution flow by processing each line to identify states. Sequential states create and merge new nodes, while decision states call \textsc{parseDecision} (Algorithm \ref{alg:parseDecision}) for further processing. 
\begin{algorithm}
\caption{Parsing Sequential Blocks}
\label{alg:parseSequential}
\begin{algorithmic}[1]
\Function{parseSequential}{startNode, lines, nodeDict}
    \State root $\gets$ startNode
    \For{each line in lines}
        \State Trim whitespace
        \If{the line represents a sequential state}
            \State Create a new node into nodeDict
            \State Merge new nodes
        \ElsIf{the line represents a decision state}
            \State Call \textsc{parseDecision}
            \State Connect new nodes
        \Else
            \State Continue processing
        \EndIf
    \EndFor
    \State \Return
\EndFunction
\end{algorithmic}
\end{algorithm}

\subsection{Parse Decision Blocks}
This algorithm handles decision structures by first parsing the "if" block using \textsc{parseSequential} (Algorithm \ref{alg:parseSequential}). If an "else if" block is encountered, it recursively calls itself to process nested conditions. For an "else" block, it parses the sequence and connects the resulting nodes using 
 \textsc{parseSequential} (Algorithm \ref{alg:parseSequential}). This ensures correct branching and flow control within decision blocks.
\begin{algorithm}
\caption{Parsing Decision Blocks}
\label{alg:parseDecision}
\begin{algorithmic}
\Function{parseDecision}{startNode, lines, nodeDict}
    \State root $\gets$ startNode
    \State Call \textsc{parseSequential}to parse ``if'' block
    \If{meet ``else if'' block}
        \State Recursively call \textsc{parseDecision}
        \State Connect new nodes
        \State \Return
    \ElsIf{meet ``else'' block}
        \State Parse the ``else'' block using \textsc{parseSequential}
        \State Connect new nodes
        \State \Return
    \Else
        \State \Return
    \EndIf
\EndFunction
\end{algorithmic}
\end{algorithm}

\begin{table*}[t]
\centering
\resizebox{\textwidth}{!}{
\begin{tabular}{|c|p{3cm}|p{3cm}|p{3cm}|p{4cm}|}
\hline
\textbf{Format} & \textbf{PlantUML} & \textbf{ChartMage} & \textbf{NomNoml} & \textbf{Meamaid} \\
\hline
\multirow{4}{*}{\textbf{Decision Block}} & 
if (D) then (C1) & D - C1 -$>>$ S1 & [D] C1-$>$ [Block1] & A\{D\} -- C1 --$>$ B[S1] \\
&\hspace*{1em} S1 & D - C2 -$>>$ S2 & [D] C2-$>$ [Block2] & A -- C2 --$>$ C[S2] \\
&else (C2) & & & \\
&\hspace*{1em} S2 & & & \\
\hline
\multirow{3}{*}{\textbf{Input}} & 
\multicolumn{4}{l|}{Here is the flowchart code:} \\
& \multicolumn{4}{l|}{[a UML flowchart with 21 paths, expressed in the corresponding format]} \\
& \multicolumn{4}{l|}{Show all possible complete paths and count how many there are.} \\
\hline
\textbf{Path Count} & 21 & 15 & 19 & 17 \\
\hline
\end{tabular}
}
\caption{Comparison of Flowchart Formats}
\label{tab:different-flowchart-formats}
\end{table*}

\section{Dataset}

\subsection{Different Format to Represent Flowchart}
\label{appentix:different-flowchart-format}

In Table \ref{tab:different-flowchart-formats}, we compares different formats to represent flowchart. We select a complex flowchart with 21 paths, and let GPT-4o to find all paths and count the number paths. Only with PlantUML, GPT-4o correctly output the right path number, which means flowchart with PlantUML is easy for models to understand. Moreover, syntax features, as is shown in Table \ref{tab:different-flowchart-formats}, such as indentation branching, and loops, making PlantUML a good format to read and convenient for program processing.

\subsection{PFDial-H Tests}
\label{appendix:pfdiag-h}
Table \ref{tab:pfd-h} presents the details of PFDial-H tests, including the total number of dialogues in the test set, average dialogue length in turns, and the proportion of dialogues with backward transition distances divided by the threshold of 5.

\section{Experimental Details}
\subsection{Implementation Details} 
\label{appendix:implementation_details}

To ensure reproducibility of our experiments, we provide detailed hyper-parameter configurations used in our training process. All experiments were conducted using the AdamW optimizer with cosine annealing learning rate scheduling. The complete set of training hyper-parameters is presented in Table \ref{tab:hyperparams}. We maintained consistent hyper-parameter settings across models of different scales to ensure fair comparison.

\subsection{Supplementary Data for Data Scaling Experiment} 
\label{appendix:scaling}
While the section \ref{sec:scale_experiment} presents the overall accuracy trend with increasing training samples, here we provide the complete experimental results. Tables \ref{tab:scaling_strategy_a} and \ref{tab:scaling_strategy_b} present the performance results for Strategy A and Strategy B respectively. Each table shows overall accuracy, decision accuracy, and sequential accuracy metrics on both ID and OOD tests across different training sample sizes. The information of dataset with different training sample sizes is shown in Table \ref{tab:scaling_datasets}.

\subsection{Details in Format Ablation Study}
\label{appendix:data and attention in different formats}

\begin{figure*}[!]
\vskip 0.2in
\begin{center}
    \includegraphics[width=0.7\textwidth]{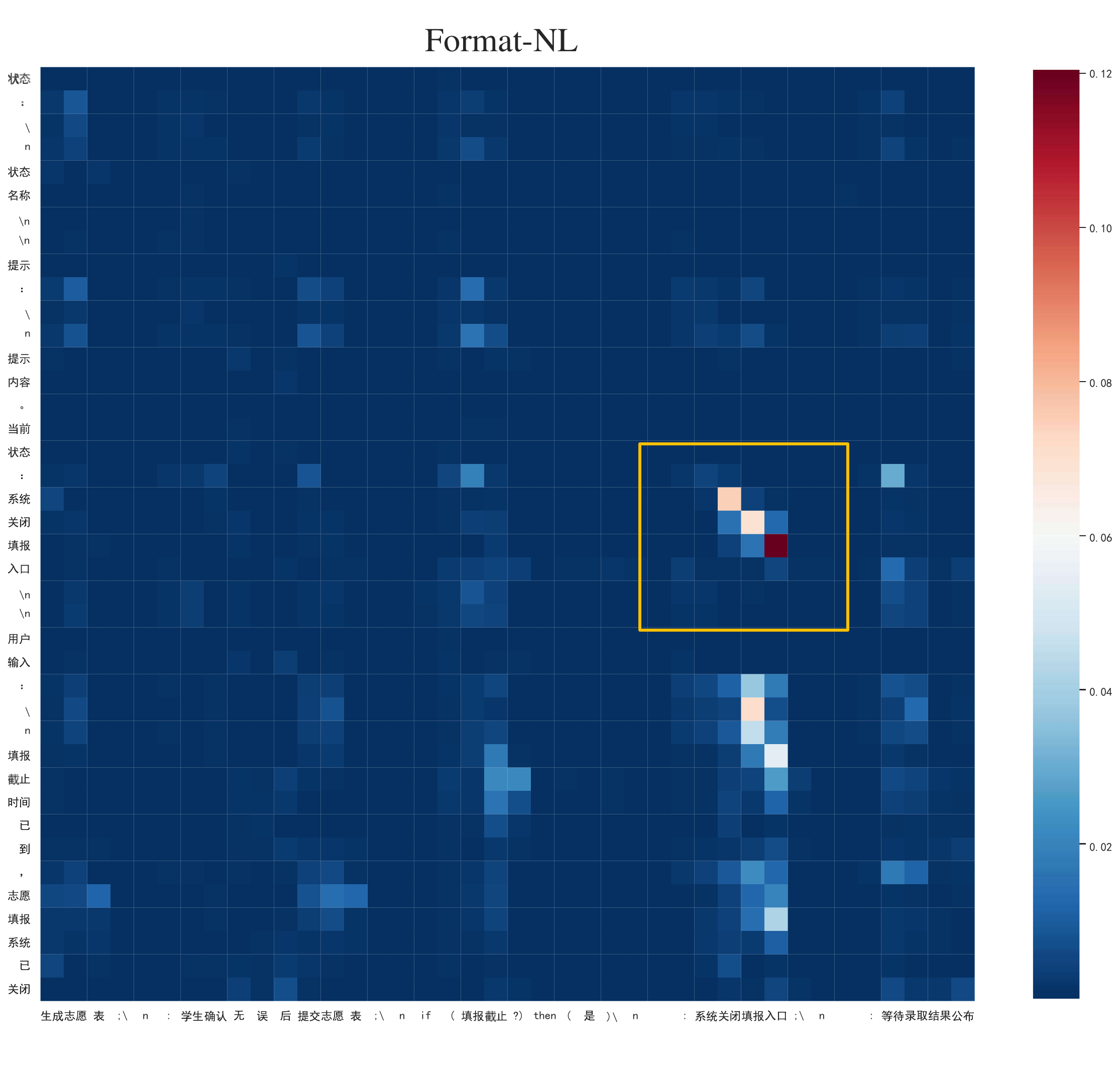} \\
    (a) Format-NL
\end{center}
\caption{local attention score of $head\_layer_{27}\_head_{20}$ using Format-NL}
\label{fig:attention_scores_a}
\vskip -0.1in
\end{figure*}

\begin{figure*}[!]
\vskip 0.2in
\begin{center}
    \includegraphics[width=0.7\textwidth]{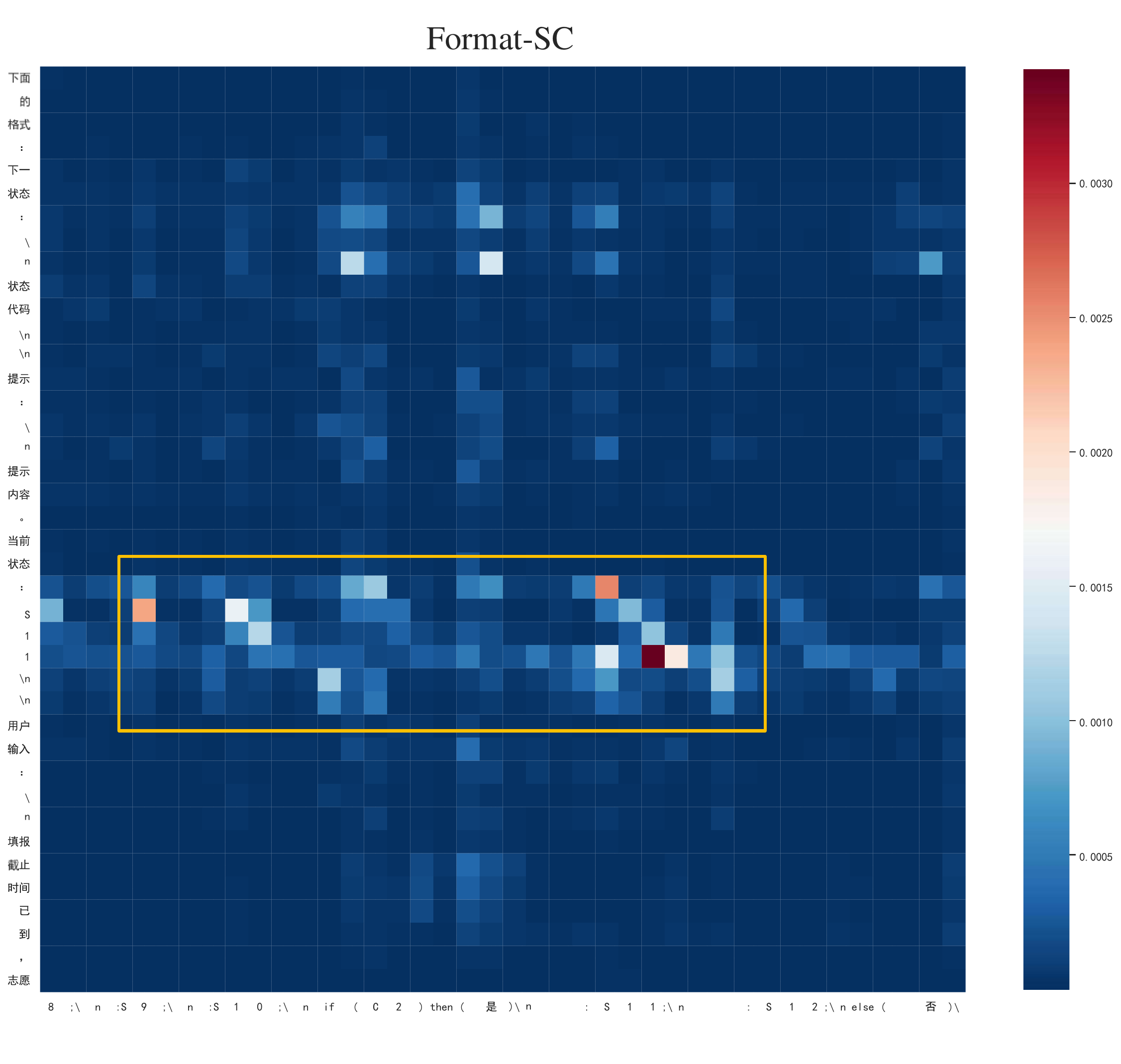} \\
    (b) Format-SC
\end{center}
\caption{local attention score of $head\_layer_{27}\_head_{20}$ using Format-SC}
\label{fig:attention_scores_b}
\vskip -0.1in
\end{figure*}

\begin{figure*}[!]
\vskip 0.2in
\begin{center}
    \includegraphics[width=0.7\textwidth]{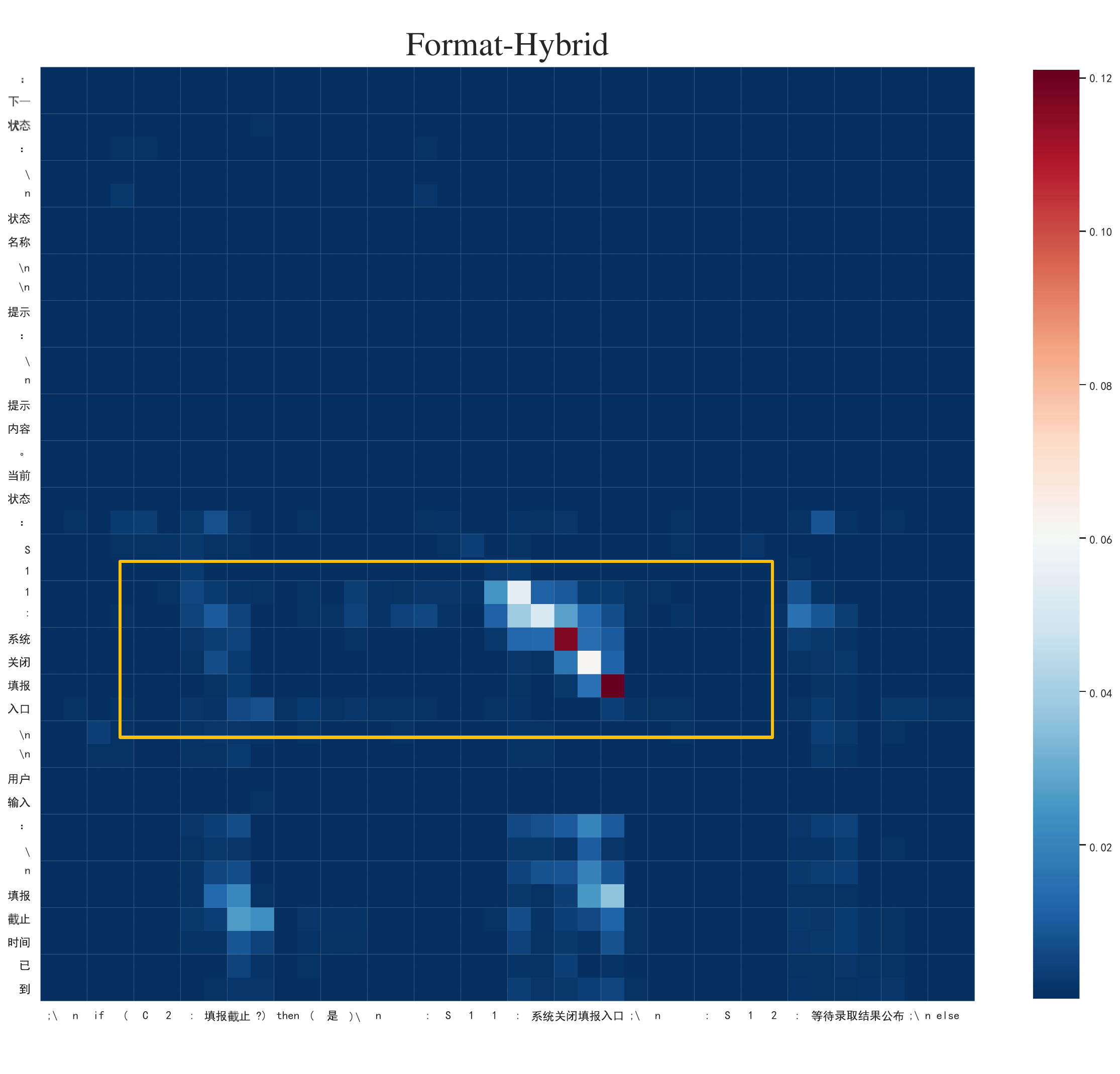} \\
    (c) Format-Hybrid
\end{center}
\caption{local attention score of $head\_layer_{27}\_head_{20}$ using Format-Hybrid}
\label{fig:attention_scores_c}
\vskip -0.1in
\end{figure*}

In this section, we present specific cases of data in different formats and the visualization results for $head\_layer_{27}\_head_{20}$. As shown in Figure \ref{fig:attention_scores_a}, \ref{fig:attention_scores_b}, and \ref{fig:attention_scores_c}, we visualize the attention patterns for Format-NL (Natural Language) in Table \ref{tab:formatA}, Format-SC (State Code) in Table \ref{tab:formatB}, and Format-Hybird in Table \ref{tab:formatC} respectively.

\begin{table}[t]
    \centering
    \begin{tabular}{|c|c|}
    \hline
    Backward Distance < 5 & 22 \\
    \hline
    Backward Distance $\geq$ 5 & 58 \\
    \hline
    Dialogue Samples & 80 \\
    \hline
    Avg. Length & 534.36 \\
    \hline
    \end{tabular}
    \caption{PFDial-H Tests Data Overview}
    \label{tab:pfd-h}
    \end{table}

\begin{table}[t]
    \centering  
    \begin{tabular}{ll}  
        \toprule  
        Hyperparameter & Value \\  
        \midrule  
        Optimizer & AdamW \\  
        Learning Rate & $5\times10^{-6}$ \\  
        Learning Rate Scheduling & Cosine Annealing \\  
        Adam Beta1 & 0.9 \\  
        Adam Beta2 & 0.95 \\  
        Batch Size & 128 \\  
        Batch Size Per-Device& 4 \\    
        Training Epochs & 5 \\     
        \bottomrule  
    \end{tabular}  
    \caption{Training Hyper Parameters}  
    \label{tab:hyperparams}  
\end{table}

\section{Prompt}
\label{appentix:Prompt}
\subsection{Prompt for generating User Input}
We use the prompt from Table \ref{tab:prompt-user-input} to generate user inputs. This prompt helps us create appropriate user input text based on the given state transitions.

\begin{table}[t]
    \centering
    \resizebox{\linewidth}{!}{
    \begin{tabular}{p{\linewidth}}
    \toprule
    \rowcolor{gray!10} \multicolumn{1}{c}{\textit{User}} \\
These examples are four-tuples consisting of the PlantUML diagram, the current state, the next state, and the user input.\\\\
\textbf{[several examples]}\\\\ 
The user's input explains the change in state from the current state to the next state. For example, if the original state is A, the user might input "A has been completed." or, when a choice is required, the user selects an option based on the next state. \\
Now I have a four-tuple consisting of the PlantUML diagram, the current state, and the next state, without user inpupt. Your task is to generate the user's input based on the rules provided. \\\\
This is the four-tuple need to be filled: \\
\textbf{[four-tuple to be filled]}\\
    \bottomrule
    \end{tabular}
    }
    \caption{The prompt for generating the user's input.}
    \label{tab:prompt-user-input}
\end{table}

\subsection{Prompt for generating Robot Output}
We use the prompt from Table \ref{tab:prompt-robot-output} to generate robot outputs. This prompt helps us create appropriate robot responses based on the current state, next state, and user input.

\begin{table}[H]
    \centering
    \resizebox{\linewidth}{!}{
    \begin{tabular}{p{\linewidth}}
    \toprule
    \rowcolor{gray!10} \multicolumn{1}{c}{\textit{User}} \\
These examples are five-tuples consisting of the PlantUML diagram, the current state, the next state, user input, and the robot output.\\\\
\textbf{[several examples]}\\\\ 
The user's input explains the change in state from the current state to the next state. The robot output is related to next state. Robot acts as the server-provider. For example, if the current state is A, tobot might output "Now 
 process A." or, when a choice is required, robot lets user to make a choice.\\
Now I have a five-tuple consisting of the PlantUML diagram, the current state next state, and the user input, without robot output. Your task is to generate the robot's output based on the rules provided. \\\\
This is the five-tuple need to be filled: \\
\textbf{[five-tuple to be filled]}\\
    \bottomrule
    \end{tabular}
    }
    \caption{The prompt for generating the robot's output.}
    \label{tab:prompt-robot-output}
\end{table}
\vspace{1cm}

\clearpage
\subsection{Prompt for Adding Backward Transition}
In the third step, we use the prompt from Table \ref{tab:prompt-backward-transition} to add backward transitions. This prompt guides us in adding logical loop structures to the flowchart.

\begin{table}[ht]
    \centering
    \resizebox{\linewidth}{!}{
    \begin{tabular}{p{\linewidth}}
    \toprule
    \rowcolor{gray!10} \multicolumn{1}{c}{\textit{User}} \\
This is a flowchart in PlantUML syntax and the result after adding a loop to itself:\\\\
\textbf{[original PlantUML and revised PlantUML]}\\\\
Your task is to follow this modification rule ro add a loop to the plantuml that I will give you next. The following conditions must be met:\\
1. The added loop is logical\\
2. The conditional state of ``repeat while'' cannot be repeated with the any conditional state that already exists in the original PlantUML\\
3. Ensure that the syntax of PlantUML is correct\\
4. Add is(\text{[need to loop]}) not(\text{[jump out of loop]}) statements after repeat while as much as possible.\\\\
PlantUML to be modified:\\
\textbf{[original PlantUML]}\\
    \bottomrule
    \end{tabular}
    }
    \caption{The prompt for Adding Backward Transition.}
    \label{tab:prompt-backward-transition}
\end{table}

\section{Case Study}
\label{case study}
We analyzed specific cases to provide intuitive explanations in Table \ref{tab:case_seq} and \ref{tab:case_dec}.
First, generally speaking, these models fail because they don't precisely provide complete descriptions of the next state, or they generate states related to user input but not in the original flowchart, or they don't follow the format required in the system prompt.
Our task examines models' ability to predict the next state in sequential branches and decision branches. Below we provide examples and analyze why GPT-4o performs poorly in these areas.

\begin{table*}
    \centering
    \resizebox{\linewidth}{!}{
    \begin{tabular}{lp{\linewidth}}
        \toprule
        PlantUML & 
        \begin{tabular}[t]{@{}l@{}}
        \texttt{@startuml} \\
        \texttt{start} \\
        \texttt{:Customer arrives at photo shop;} \\
        \texttt{:Submit photo files for printing;} \\
        \texttt{:Select print size and quantity;} \\
        \texttt{:Choose paper quality and surface effect;} \\
        \texttt{:Confirm print order and price;} \\
        \texttt{if (Photo quality check passed?) then (yes)} \\
        \hspace*{2em}\texttt{:Pay fee;} \\
        \hspace*{2em}\texttt{:Photo printing in progress;} \\
        \hspace*{2em}\texttt{if (Printing completed?) then (yes)} \\
        \hspace*{4em}\texttt{:Display printed photos;} \\
        \hspace*{4em}\texttt{:Customer satisfied?;} \\
        \hspace*{4em}\texttt{if (Customer satisfied?) then (yes)} \\
        \hspace*{6em}\texttt{:Complete transaction;} \\
        \hspace*{4em}\texttt{else (no)} \\
        \hspace*{6em}\texttt{:Reselect photos;} \\
        \hspace*{6em}\texttt{:Reprint;} \\
        \hspace*{6em}\texttt{:Complete transaction;} \\
        \hspace*{4em}\texttt{endif} \\
        \hspace*{2em}\texttt{else (no)} \\
        \hspace*{4em}\texttt{:Wait for photo printing;} \\
        \hspace*{4em}\texttt{:Repeatedly check if printing is complete;} \\
        \hspace*{2em}\texttt{endif} \\
        \texttt{else (no)} \\
        \hspace*{2em}\texttt{:Notify poor photo quality;} \\
        \hspace*{2em}\texttt{:Reselect photos;} \\
        \hspace*{2em}\texttt{:Reprint;} \\
        \hspace*{2em}\texttt{:Complete transaction;} \\
        \texttt{endif} \\
        \texttt{:Customer leaves the photo shop;} \\
        \texttt{stop} \\
        \texttt{@enduml} \\
        \end{tabular} \\
        \midrule
        Current State & Repeatedly check if printing is complete \\
        \midrule
        User Input & Printing is completed. \\
        \midrule
        Predicted Next State & \textcolor{red}{\texttt{Display printed photos} (Incorrect)} \\
        \midrule
        Correct Next State & \texttt{Customer leaves the photo shop} (Correct) \\
        \bottomrule
    \end{tabular}
    }
    \caption{Case Study in Sequential Branch}
    \label{tab:case_seq}
\end{table*}
\subsection{Sequential Branch}
For sequential branches, the model only need to capture the current state and state transition. This is typically simple because adjacent states in PlantUML are usually adjacent lines. However, when multiple decision branches are nested, things become complicated and the line distance between adjacent states increases.
For example, in Table \ref{tab:case_seq}, after "Repeatedly check if printing is complete", all nested loops have ended and should transition to "Customer leaves the photo shop". However, due to the high complexity caused by multiple nested sequential branches, GPT-4o struggles to accurately capture the contextual state relationships, resulting in errors.

\subsection{Decision Branch}
For decision branches, the model needs to precisely transition to the next state based on the current state and user input. Unfortunately, in most wrong cases, due to the high similarity between user input and the first state of the correct branch during condition judgment, the model tends to skip this first state and directly enter subsequent states, failing to accurately follow the agreed path. More challenging issues arise when decision branch conditions involve more than just yes/no decisions, requiring models to comprehensively understand user input and judge their true intentions.
For example, in Table \ref{tab:case_dec}, when the user inputs "The position of the lighting fixture needs to be adjusted.", it indicates that the lighting position needs to be modified at this point. However, the model struggles to strictly adhere to the process constraints and makes an incorrect state prediction based on the user input, skipping the correct state of this branch.

\begin{table*}
    \centering
    \resizebox{\linewidth}{!}{
    \begin{tabular}{lp{\linewidth}}
        \toprule
        PlantUML & 
        \begin{tabular}[t]{@{}l@{}}
        \texttt{@startuml} \\
        \texttt{start} \\
        \texttt{:Communicate lighting design plan with construction team;} \\
        \texttt{:Explain design requirements and fixture layout;} \\
        \texttt{:Provide technical support and guidance;} \\
        \texttt{:Coordinate installation time and progress;} \\
        \texttt{:Supervise fixture installation process;} \\
        \texttt{:Perform fixture debugging and brightness adjustment;} \\
        \texttt{repeat} \\
        \texttt{:Confirm fixture layout and installation position;} \\
        \texttt{:Supervise fixture installation process;} \\
        \texttt{repeat while(Installation and debugging unsuccessful)} \\
        \texttt{:Check fixture installation quality and safety;} \\
        \texttt{if (Need to adjust fixture position?) then (yes)} \\
        \hspace*{2em}\texttt{:Negotiate adjustment plan;} \\
        \hspace*{2em}\texttt{:Reinstall or adjust fixture position;} \\
        \texttt{else (no)} \\
        \hspace*{2em}\texttt{:Confirm final installation result;} \\
        \texttt{endif} \\
        \texttt{:Final acceptance of lighting system;} \\
        \texttt{:Ensure lighting system meets design requirements;} \\
        \texttt{:Complete construction documents and records;} \\
        \texttt{stop} \\
        \texttt{@enduml} \\
        \end{tabular} \\
        \midrule
        Current State & Need to adjust fixture position? \\
        \midrule
        User Input & The position of the lighting fixture needs to be adjusted. \\
        \midrule
        Predicted Next State & \textcolor{red}{\texttt{Reinstall or adjust fixture position} (Incorrect)} \\
        \midrule
        Correct Next State & \texttt{Negotiate adjustment plan} (Correct) \\
        \bottomrule
    \end{tabular}
    }
    \caption{Case Study in Decision Branch}
    \label{tab:case_dec}
\end{table*}

\clearpage
\newpage
\begin{table*}
  \small
  \centering
  \setlength{\tabcolsep}{6pt} 
  \renewcommand{\arraystretch}{1.1} 
  \begin{tabular}{lrrrrrrrrr}  
    \toprule
    \textbf{Training Sample Size} & \textbf{100} & \textbf{200} & \textbf{400} & \textbf{800} & \textbf{1600} & \textbf{3200} & \textbf{6400} & \textbf{9600} & \textbf{12705} \\
    \midrule
    Flowcharts           & 3   & 5   & 12  & 29  & 59   & 113  & 208  & 308  & 440  \\
    State Nodes         & 49  & 79  & 149 & 304 & 612  & 1236 & 2473 & 3772 & 5055 \\
    Sequential Samples  & 62  & 134 & 264 & 561 & 1099 & 2185 & 4293 & 6717 & 9029 \\
    Decision Samples    & 38  & 66  & 136 & 239 & 501  & 1015 & 2107 & 2883 & 3676 \\
    Avg. Length        & 386.22 & 368.19 & 320.24 & 286.92 & 275.63 & 272.29 & 273.00 & 280.90 & 277.16 \\
    \bottomrule
  \end{tabular}
  \caption{Statistics of the PFDial Dataset with Different Training Sample Sizes.}
  \label{tab:scaling_datasets}
\end{table*}

\begin{table*}
    \small
    \centering
    \newcolumntype{L}{>{\raggedright\arraybackslash}p{3.5cm}} 
    \newcolumntype{C}{>{\centering\arraybackslash}p{1.5cm}} 
    \begin{tabular}{L|C C C|C C C}  
    \toprule
    \multirow{2}{*}{\parbox{3.5cm}{\centering \textbf{Training Sample Size}}} & \multicolumn{3}{c|}{\textbf{ID-test}} & \multicolumn{3}{c}{\textbf{OOD-test}} \\
    \cmidrule(lr){2-4} \cmidrule(lr){5-7}
    & \textbf{Acc} & \textbf{Decision Acc} & \textbf{Sequential Acc} 
    & \textbf{Acc} & \textbf{Decision Acc} & \textbf{Sequential Acc} \\
    \midrule
    100-all & $77.61$ & $54.52$ & $81.87$ & $73.59$ & $39.81$ & $79.08$ \\
    200-all & $84.82$ & $83.33$ & $85.10$ & $80.83$ & $62.35$ & $83.83$ \\
    400-all & $88.91$ & $90.68$ & $88.59$ & $87.80$ & $76.50$ & $89.64$ \\
    800-all & $90.37$ & $92.37$ & $89.99$ & $88.44$ & $78.90$ & $89.99$ \\
    1600-all & $93.84$ & $94.92$ & $93.64$ & $91.59$ & $84.17$ & $92.79$ \\
    3200-all & $96.61$ & $97.46$ & $96.46$ & $93.26$ & $89.69$ & $93.84$ \\
    6400-all & $97.89$ & $97.18$ & $98.02$ & $95.34$ & $91.85$ & $95.91$ \\
    9600-all & $98.86$ & $\mathbf{98.31}$ & $98.96$ & $96.45$ & $90.17$ & $\mathbf{97.47}$ \\
    all & $\mathbf{98.94}$ & $98.02$ & $\mathbf{99.11}$ & $\mathbf{96.51}$ & $\mathbf{90.65}$ & $\mathbf{97.47}$ \\
    \bottomrule
    \end{tabular}
    \caption{Performance with different training sample sizes across ID and OOD datasets after training on Qwen2.5-7B, with data scaling strategy \textbf{(a)}.}
    \label{tab:scaling_strategy_a}
\end{table*}

\begin{table*}
    \small
    \centering
    \newcolumntype{L}{>{\raggedright\arraybackslash}p{3.5cm}} 
    \newcolumntype{C}{>{\centering\arraybackslash}p{1.5cm}} 
    \begin{tabular}{L|C C C|C C C}  
    \toprule
    \multirow{2}{*}{\parbox{3.5cm}{\centering \textbf{Training Sample Size}}} & \multicolumn{3}{c|}{\textbf{ID-test}} & \multicolumn{3}{c}{\textbf{OOD-test}} \\
    \cmidrule(lr){2-4} \cmidrule(lr){5-7}
    & \textbf{Acc} & \textbf{Decision Acc} & \textbf{Sequential Acc} 
    & \textbf{Acc} & \textbf{Decision Acc} & \textbf{Sequential Acc} \\
    \midrule
    100-100 & $61.37$ & $23.45$ & $68.37$ & $59.12$ & $13.91$ & $66.46$ \\ 
    200-200 & $82.31$ & $84.46$ & $81.92$ & $77.88$ & $65.47$ & $79.90$ \\ 
    400-400 & $89.71$ & $89.55$ & $89.73$ & $86.03$ & $75.30$ & $87.77$ \\ 
    800-800 & $91.55$ & $92.66$ & $91.35$ & $90.01$ & $81.06$ & $91.47$ \\ 
    1600-1600 & $93.80$ & $93.50$ & $93.85$ & $91.79$ & $85.13$ & $92.87$ \\ 
    3200-3200 & $97.01$ & $98.02$ & $96.82$ & $94.30$ & $90.65$ & $94.90$ \\ 
    6400-6400 & $97.98$ & $98.31$ & $97.92$ & $95.54$ & $90.65$ & $96.34$ \\ 
    9600-9600 & $98.90$ & $97.46$ & $\mathbf{99.17}$ & $95.95$ & $89.45$ & $97.00$ \\ 
    all & $\mathbf{98.94}$ & $\mathbf{98.02}$ & ${99.11}$ & $\mathbf{96.51}$ & $\mathbf{90.65}$ & $\mathbf{97.47}$ \\
    \bottomrule
    \end{tabular}
    \caption{Performance with different training sample sizes across ID and OOD datasets after training on Qwen2.5-7B, with data scaling strategy \textbf{(b)}.}
    \label{tab:scaling_strategy_b}
\end{table*}

\newpage

\newpage

\clearpage
\begin{table*}
    \centering
    \resizebox{\linewidth}{!}{
    \begin{tabular}{lp{\linewidth}}
        \toprule
        Format & Data Example \\
        \midrule
        Format-NL & \textbf{PlantUML:}\\
        &@startuml start :College entrance exam results announced; :Student obtains college entrance exam score report; :Student checks the exam results and determines the range of colleges and majors to apply for; if (Do you need to research colleges and majors in advance?) then (Yes) :Student conducts research on colleges and majors; else (No) :Skip this step; endif :Student logs into the application system; :System provides the application entry and instructions; :Student fills out the application, prioritizing choices; :After completing the application, the system generates an application form; :Student confirms the form and submits it; if (Application deadline?) then (Yes) :System closes the application entry; :Wait for the admission results to be announced; else (No) :Student can modify the application before the deadline; :Wait for the application deadline; endif :Admission results announced; if (Admitted?) then (Yes) :Student completes registration procedures according to the admission notice; else (No) :Student applies for supplementary applications or participates in the supplementary application process; endif :Enroll in the school; stop @enduml \\
        &\textbf{Current state:} \\
        &System closes the application entry \\
        &\textbf{Next state:} \\
        &Wait for the admission results to be announced \\
        &\textbf{User input:} \\
        &The application deadline has passed, and the application system is now closed. \\
        &\textbf{Robot output:} \\
        &Please patiently wait for the announcement of the admission results. \\
        \bottomrule
    \end{tabular}
    }
    \caption{Example Data Display - Format-NL}
    \label{tab:formatA}
\end{table*}

\clearpage
\begin{table*}
    \centering
    \resizebox{\linewidth}{!}{
    \begin{tabular}{lp{\linewidth}}
        \toprule
        Format & Data Example \\
        \midrule
        Format-SC & \textbf{PlantUML:} \\
        &@startuml
start
:S1;
:S2;
:S3;
if (C1) then (Yes)
:S4;
else (No)
:S5;
endif
:S6;
:S7;
:S8;
:S9;
:S10;
if (C2) then (Yes)
:S11;
:S12;
else (No)
:S13;
:S14;
endif
:S15;
if (C3) then (Yes)
:S16;
else (No)
:S17;
endif
:S18;
stop
@enduml \\
        &\textbf{Dictionary of the state codes:} \\
        &\{
"College entrance exam results announced": "S1",
"Student obtains college entrance exam score report": "S2",
"Student checks the exam results and determines the range of colleges and majors to apply for": "S3",
"Do you need to research colleges and majors in advance?": "C1",
"Student conducts research on colleges and majors": "S4",
"Skip this step": "S5",
"Student logs into the application system": "S6",
"System provides the application entry and instructions": "S7",
"Student fills out the application, prioritizing choices": "S8",
"After completing the application, the system generates an application form": "S9",
"Student confirms the form and submits it": "S10",
"Application deadline?": "C2",
"System closes the application entry": "S11",
"Wait for the admission results to be announced": "S12",
"Student can modify the application before the deadline": "S13",
"Wait for the application deadline": "S14",
"Admission results announced": "S15",
"Admitted?": "C3",
"Student completes registration procedures according to the admission notice": "S16",
"Student applies for supplementary applications or participates in the supplementary application process": "S17",
"Enroll in the school": "S18"
\} \\
        &\textbf{Current state:} \\
        &System closes the application entry \\
        &\textbf{Next state:}  \\
        &Wait for the admission results to be announced \\
        &\textbf{User input:}  \\
        &The application deadline has passed, and the application system is now closed.  \\
        &\textbf{Robot output:}   \\
        &Please patiently wait for the announcement of the admission results. \\
        \bottomrule
    \end{tabular}
    }
    \caption{Example Data Display - Format-SC} 
    \label{tab:formatB}
\end{table*}

\clearpage
\begin{table*}
    \centering
    \resizebox{\linewidth}{!}{
    \begin{tabular}{lp{\linewidth}}
        \toprule
        Format & Data Example \\
        \midrule
        Format-Hybrid & \textbf{PlantUML:} \\
        &@startuml
start
:S1: College entrance exam results announced;
:S2: Student obtains college entrance exam score report;
:S3: Student checks the exam results and determines the range of colleges and majors to apply for;
if (C1: Do you need to research colleges and majors in advance?) then (Yes)
:S4: Student conducts research on colleges and majors;
else (No)
:S5: Skip this step;
endif
:S6: Student logs into the application system;
:S7: System provides the application entry and instructions;
:S8: Student fills out the application, prioritizing choices;
:S9: After completing the application, the system generates an application form;
:S10: Student confirms the form and submits it;
if (C2: Application deadline?) then (Yes)
:S11: System closes the application entry;
:S12: Wait for the admission results to be announced;
else (No)
:S13: Student can modify the application before the deadline;
:S14: Wait for the application deadline;
endif
:S15: Admission results announced;
if (C3: Admitted?) then (Yes)
:S16: Student completes registration procedures according to the admission notice;
else (No)
:S17: Student applies for supplementary applications or participates in the supplementary application process;
endif
:S18: Enroll in the school;
stop
@enduml \\
        &\textbf{Current state:} \\
        &System closes the application entry \\
        &\textbf{Next state:} \\
        &Wait for the admission results to be announced \\
        &\textbf{User input:} \\
        &The application deadline has passed, and the application system is now closed. \\
        &\textbf{Robot output:} \\
        &Please patiently wait for the announcement of the admission results. \\
        \bottomrule
    \end{tabular}
    }
    \caption{Example Data Display - Format-Hybrid} 
    \label{tab:formatC}
\end{table*}

\clearpage
\begin{table*}
    \centering
    \resizebox{\linewidth}{!}{
    \begin{tabular}{p{4cm}p{10cm}}
        \toprule
        \textbf{Category} & \textbf{Senarios} \\
        \midrule
        Lifestyle Services & Hairdressing, Phone Card, Car Wash, Agritainment, Printing \& Copying, Lawyer, Yoga Studio, Music Classroom, Internship \\
        \midrule
        Daily Convenience & Takeout, Scenic Spots, Furniture Cleaning, Wedding Photography, Movie, Cleaning Service, Self-service Car Wash, Second-hand Car Trading, Visa Application \\
        \midrule
        Food \& Education & Catering, Training Courses, Physiotherapy \& Massage, Lighting Design, Gym, Express Delivery, Travel Group Purchase, Health Check-up, Group Tour \\
        \midrule
        Entertainment \& Transportation & Concert, Car Rental, Baking Studio, Digital Repair, Airplane Ticket, Water Delivery, Florist Shop, Photo Studio, Course Selection \\
        \midrule
        Business \& Professional Services & Commercial Photography, Hospital, Pet Boarding, Café, Dentist, Pet Grooming, Tea House, Outdoor Expansion, Electrician Inspection \\
        \midrule
        Luxury \& Specialized Services & Beauty Salon, Museum, Horticultural Design, Car Maintenance, Cruise, Photography Studio Rental, Piano Tuning, Basketball Court, Cargo Delivery into Cabin \\
        \midrule
        Living-related Services & Laundry, House Rental, Education Consultation, Library, Cultural Exhibition, Health Consultation, Holiday Villa, Interior Design, Bank Account Operation \\
        \midrule
        Maintenance \& Care Services & Home Appliance Repair, Hotel, Leather Goods Care, Arcade, Furniture Installation, Medical Check-up, Car Insurance Claim, Bicycle Rental, Parts Inspection \\
        \midrule
        Comprehensive Services & Moving, Resort, Language Translation, Medical Aesthetics, Driving School, Wedding Planning, Pet Hospital, Manicure, Document Approval \\
        \midrule
        Shopping \& Other Activities & Shopping, Train Ticket, Water \& Electricity Repair, Ski Resort, Credit Card, College Entrance Examination Volunteer Filling, Cooking, DIY Handicraft, Content Creation \\
        \bottomrule
    \end{tabular}
    }
    \caption{Senarios}
    \label{tab:Business Scenarios}
\end{table*}

\end{document}